\documentclass{article} 
\usepackage[final]{colm2026_conference}

\usepackage{microtype}
\usepackage{hyperref}
\usepackage{url}
\usepackage{booktabs}
\usepackage{amsmath,amssymb}
\usepackage{graphicx}
\usepackage{multirow}
\usepackage{makecell}
\usepackage{enumitem}
\usepackage{xcolor}
\usepackage{colortbl}
\usepackage{algorithm}
\usepackage{algorithmic}
\usepackage{pifont}
\usepackage{float}
\usepackage{tcolorbox}
\tcbuselibrary{listings,skins,breakable}

\newtcolorbox{promptbox}[1][]{
       enhanced, breakable,
       colback=gray!5, colframe=gray!60,
       fonttitle=\small\bfseries\sffamily,
       coltitle=white, colbacktitle=gray!70,
       boxrule=0.5pt, arc=2pt,
       left=6pt, right=6pt, top=4pt, bottom=4pt,
       title={#1}
}

\usepackage{lineno}

\definecolor{darkblue}{rgb}{0, 0, 0.5}
\hypersetup{colorlinks=true, citecolor=darkblue, linkcolor=darkblue, urlcolor=darkblue, hypertexnames=false}

\title{Liberating LLM Capabilities in Full-Duplex Speech Models}

\author{Luoyuan Zhang$^{1,2}$\thanks{This work was done while Luoyuan Zhang was an intern at ModelBest.}, Bokai Xu$^{2}$, Junbo Cui$^{2}$, Weiyue Sun$^{2}$,\\ \bfseries Yingjing Xu$^{2}$, Hanyu Liu$^{2,4}$, Yuan Yao$^{2,3}$\thanks{Corresponding author. Email: \texttt{yaoyuanthu@mail.tsinghua.edu.cn}.}\\
$^{1}$CUHK(SZ) \quad $^{2}$ModelBest \quad $^{3}$Tsinghua University \quad $^{4}$Peking University
}

\newcommand{\tablestyle}{%
       \footnotesize
       \renewcommand{\arraystretch}{1.15}
       \setlength{\tabcolsep}{5pt}%
}
\newcommand{\thdr}[1]{{\scriptsize\bfseries #1}}
\newcommand{\theadcell}[1]{\makecell[c]{\scriptsize\bfseries\boldmath #1}}

\begin{document}

\ifcolmsubmission
       \linenumbers
\fi

\maketitle


\begin{abstract}
       Speech-based large language models are typically constrained to spoken replies, which limits their user-facing outputs to what can be verbalized and suppresses text-native capabilities such as code generation, structured analysis, and multi-step reasoning in realtime interaction, for tasks that require persistent, structured, and inspectable intermediate outputs. Existing work improves spoken reasoning or full-duplex turn-taking, but still treats text as a hidden intermediate state or a subordinate modality rather than a first-class output channel. We propose \textbf{Listen-Write-Speak (LWS)}, a text-first tri-channel paradigm in which a single autoregressive LLM continuously \emph{listens} to user audio, \emph{writes} visible free-form text as its primary output, and \emph{speaks} a realtime oral response in parallel under a shared causal attention context. This behavior is implemented entirely through a \emph{Token Schema}, requiring no architectural modifications, and learned via a two-stage data pipeline that synthesizes per-second cognitive annotations consistent with the revealed input timeline. Empirically, LWS demonstrates strong full-duplex interaction on Full-Duplex-Bench, reaches 4.72 on VoiceBench AlpacaEval, achieves 92.6\% writing and speaking consistency, and consistently outperforms its internal ablations on URO-Bench. These results suggest that visible writing can serve as a first-class output channel for speech interaction while preserving low-latency use. The code, dataset, and demo are available on the project page: \url{https://royalzhang.com/project/lws-page/}.
\end{abstract}

\section{Introduction}

Human communication is inherently multimodal, yet speech and writing do not share the burden equally. Decades of work in cognitive science, CSCW, and multimodal HCI suggest that different representational media support different functions: speech is especially effective for turn-taking, grounding, and pragmatic coordination, whereas written or visual representations are better suited to precise, structured, persistent, and spatially organized information~\citep{clark1991grounding,oviatt1999ten,larkin1987diagram,scaife1996external}. In practice, the two modalities are almost always co-present: collaborative software engineering revolves around shared source files that participants discuss verbally; scientific presentations pair spoken narrative with projected figures and equations; and professional meetings produce written minutes that outlive the spoken exchange. The underlying principle is consistent: speech orchestrates the interaction, while anything that must be exact, revisitable, or spatially organized is \emph{written down and laid out}.

Large language models are, fundamentally, text-native systems that share this affinity for writing. Many of their most distinctive capabilities, including code generation, structured analysis, mathematical derivation, and professional writing, are defined and most naturally expressed within the unconstrained space of written text. However, when these models interact with users through speech, a fundamental bottleneck arises: their outputs must be projected into the much narrower subspace of \emph{speakable} language. If a developer says, ``write me a binary search in Python,'' a text-based LLM can return clean, executable code; a speech-based LLM, by contrast, can only \emph{dictate} that code aloud, forcing the user to transcribe each token manually. Likewise, when a meeting participant asks, ``summarize the key decisions,'' the ideal response may be a structured Markdown table, yet a speech interface must linearize it into a spoken list that is difficult to revise. In both scenarios, the speech modality suppresses precisely those forms of output that make LLMs uniquely useful, which are also the forms that human collaborators would rarely entrust to speech alone.

A growing body of work has sought to mitigate this limitation by introducing \emph{reasoning} into speech-based LLMs in several ways: before speaking~\citep{wu2025step,woo2025think}, interleaved with speaking~\citep{chiang2025stitch,xie2025mini}, during listening~\citep{chiang2025shanks,wu2025chronological}, or through full-duplex designs~\citep{defossez2024moshi,ma2025language}. Despite their diversity, these approaches share a common premise: text remains in the role of supporting speech. In other words, reasoning is used as an internal or auxiliary process to improve the spoken response, rather than as \emph{writing}: a first-class, user-visible, editable output channel through which the model directly expresses its full capabilities.

\begin{figure*}[t]
       \centering
       \includegraphics[width=\textwidth]{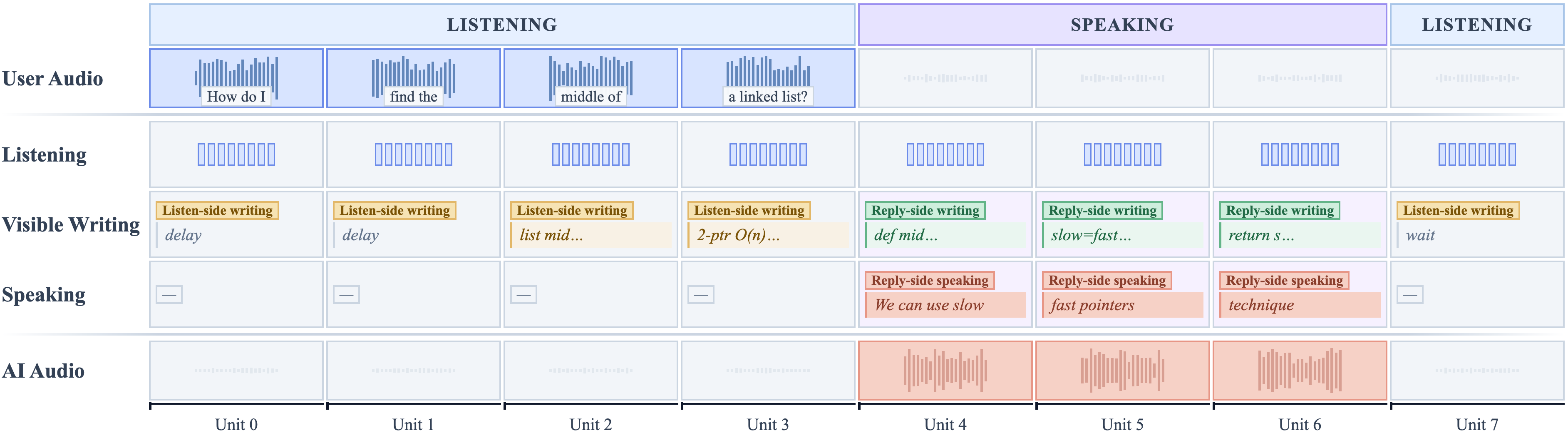}
       \caption{Listen-Write-Speak (LWS) tri-channel architecture. The interaction is partitioned into temporal Units, within which three channels operate concurrently under a shared causal context: \textbf{Listening}, \textbf{Visible Writing}, and \textbf{Speaking}. The writing channel remains active across both phases, appearing as \textit{listening-side writing} during user speech and as \textit{reply-side writing} during response, while speech is emitted only in speaking Units.}
       \label{fig:architecture}
\end{figure*}

We propose \textbf{Listen-Write-Speak (LWS)}, a paradigm that \textbf{reverses the usual hierarchy of speech interfaces}: the model still listens continuously, but it \textbf{writes first} and speaks in parallel. As illustrated in Figure~\ref{fig:architecture}, the interaction is divided into discrete temporal \emph{Units}, and each Unit is interpreted through three concurrent channels under a \textbf{shared causal context}: (1)~\textbf{Listening}, which continuously ingests the user's audio stream; (2)~\textbf{Visible Writing}, which emits free-form text that is directly shown to the user and is \textbf{not constrained to be speakable}; and (3)~\textbf{Speaking}, which verbalizes an oral response in real time. Crucially, the writing channel remains active across \textbf{both} phases of interaction: during user speech it appears as \textit{listening-side writing}, allowing the model to externalize incremental understanding before the user has finished; during response generation it becomes \textit{reply-side writing}, producing structured artifacts such as code, lists, or derivations while speech continues. By contrast, the speaking channel is active only in speaking Units, making speech a synchronized \emph{companion} to visible text rather than the sole final output. Technically, this yields a \textbf{tri-channel visible-writing architecture} inside a standard autoregressive LLM, exploiting a simple asymmetry: humans naturally speak, while LLMs are strongest when they write. Because the entire behavior is specified through token-level structure (a \emph{Token Schema}) with no new parameters, the model needs to learn only a new interaction protocol on top of capabilities it already has. Our contributions are as follows: (1)~We propose \textbf{Listen-Write-Speak (LWS)}, a text-first tri-channel paradigm in which the model continuously listens to the user's audio stream, writes visible text as its primary output, and speaks a realtime oral response in parallel, during both listening and response generation, all within a single autoregressive LLM sharing one causal attention context (Table~\ref{tab:paradigm}). (2)~We introduce a \textbf{Token Schema} that encodes the tri-channel behavior entirely through special tokens within a standard autoregressive Transformer, requiring no architectural modifications. (3)~We design a \textbf{two-stage data pipeline} that synthesizes per-second annotations intended to be causally consistent with the revealed input timeline. (4)~Experiments show that LWS demonstrates strong abilities on Full-Duplex-Bench, reaches 4.72 on VoiceBench AlpacaEval, achieves 92.6\% writing and speaking consistency, and consistently outperforms its internal ablations on URO-Bench.

\section{Related Work}
\label{sec:related}
The works most related to ours fall into three categories: reasoning-augmented speech LLMs, full-duplex spoken interaction, and multi-channel output architectures. These directions address complementary aspects of the problem and are often evaluated under different criteria. Table~\ref{tab:paradigm} summarizes them along four capability dimensions relevant to our setting.
\begin{table}[t]
       \centering
       
       \tablestyle
       \begin{tabular*}{\columnwidth}{@{\extracolsep{\fill}}p{4.2cm}p{6.1cm}cccc@{}}
              \toprule
              \textbf{Paradigm} & \textbf{models}                    & \textbf{FD} & \textbf{FT} & \textbf{CL} & \textbf{CS} \\
              \midrule
              Think-Before-Speak
              & Step-Audio~2, TVS
              & \ding{55}                           & \ding{51}   & \ding{55}   & \ding{55}                 \\
              Interleaved Think-Speak
              & STITCH, Mini-Omni-Reasoner
              & \ding{55}                           & \ding{51}   & \ding{55}   & \ding{51}                 \\
              Dual-Brain Think-Speak
              & MPS
              & \ding{55}                           & \ding{51}   & \ding{55}   & \ding{51}                 \\
              Think-While-Listen
              & SHANKS, Chron.\ Think.
              & \ding{51}$^\dagger$                 & \ding{51}   & \ding{51}   & \ding{55}                 \\
              Full-Duplex
              & Moshi, LSLM,  FlexDuo
              & \ding{51}                           & \ding{55}   & \ding{55}   & \ding{55}                 \\
              Parallel Text-Speech
              & Qwen3-Omni, Kimi-Audio
              & \ding{55}                           & \ding{51}   & \ding{55}   & \ding{55}                 \\
              Multi-Channel Protocol
              & OpenAI Harmony
              & \ding{55}                           & \ding{51}   & \ding{55}   & \ding{55}                 \\
              \midrule
              \textbf{Listen-Write-Speak}
              & \textbf{Ours}
              & \ding{51}                           & \ding{51}   & \ding{51}   & \ding{51}                 \\
              \bottomrule
       \end{tabular*}

       \vspace{2pt}
       {\footnotesize $^\dagger$Chronological Thinking only; others are half-duplex.}
\caption{Comparison of speech-LLM paradigms. \textbf{FD}: full duplex interaction; \textbf{FT}: free text outputs; \textbf{CL}: cognition while listening; \textbf{CS}: cognition while speaking. Prior models' cognition is hidden internal thinking; LWS exposes it as visible text output.}
       \label{tab:paradigm}
\end{table}

\subsection{Reasoning-augmented speech LLMs}
Recent speech LLMs incorporate explicit reasoning, but differ in when the reasoning is performed. Think-Before-Speak models such as Step-Audio~2, Step-Audio-R1, and TVS~\citep{wu2025step,tian2025step,woo2025think} reason before producing speech; this can improve response quality, but it increases latency and provides no cognition during the user's turn. Interleaved Think-Speak models such as STITCH and Mini-Omni-Reasoner~\citep{chiang2025stitch,xie2025mini}, as well as dual-model designs such as MPS~\citep{wu2025mind}, move part of the reasoning process into reply generation, but still couple cognition to the speaking phase rather than to a persistent user-visible channel. Think-While-Listen models such as SHANKS and Chronological Thinking~\citep{chiang2025shanks,wu2025chronological,shih2025can} perform reasoning during user speech, but this reasoning is introduced to support subsequent speech generation rather than as a persistent, user-visible, free-form reasoning process; accordingly, the explicit cognition ceases once reply generation begins.

\subsection{Full-duplex speech interaction}
Another line of work studies full-duplex spoken interaction. Models such as Moshi, LSLM, and FlexDuo~\citep{defossez2024moshi,ma2025language,liao2025flexduo} emphasize overlap handling, turn-taking, and low-latency response. They can listen while speaking, but typically expose only speech as the user-facing output modality. Cascaded pipelines built from ASR, an LLM, and TTS preserve strong text generation quality, but they usually revert to turn-based interaction and therefore do not provide genuine full-duplex behavior.

\subsection{Multi-channel output architectures}
Recent multimodal architectures show that speech systems can emit multiple concurrent streams. Parallel text-speech models such as Qwen3-Omni, DrVoice, and Kimi-Audio~\citep{xu2025qwen3,tan2025drvoice,ding2025kimi} generate text alongside speech, and protocol-based interfaces such as OpenAI Harmony~\citep{openai-harmony} expose additional structured channels. LWS does not claim the first multi-channel schema; its contribution is applying such a protocol to full-duplex speech with always-on visible writing. Table~\ref{tab:paradigm} summarizes how prior systems generally sacrifice at least one of four capabilities relevant to our setting: full-duplex interaction, free-text output, cognition while listening, or cognition while speaking. Our formulation differs in treating visible writing as a first-class channel that remains active across both listening and speaking.

\section{Method}
\label{sec:method}

\subsection{Architecture overview}
\label{sec:arch}

The gap identified in Related Work suggests a common limitation of prior speech LLMs: they typically preserve only part of the functionality required for rich real-time interaction, and many rely on additional architectural machinery that makes the resulting systems harder to scale and reuse. Our starting point is that these four target capabilities need not be implemented as four separate mechanisms. Instead, they can be organized as three concurrent channels that run within one full-duplex autoregressive process: Listening continuously processes the user's incoming audio, Visible Writing externalizes the model's text-native output as user-visible free-form text, and Speaking verbalizes a realtime oral response. Under this view, an always-on Listening channel provides full-duplex interaction; an always-on Visible Writing channel provides unconstrained text output and can already serve as an externalized working memory while the user is still speaking; and the joint activation of Writing and Speaking during response generation yields cognition while speaking. This design is realized in a standard autoregressive Transformer without introducing extra decoders or cross-channel fusion modules. The system operates entirely at the \emph{text-token level}: user speech is represented as discrete audio tokens, the model emits speech tokens for the Speaking channel, and their hidden states are passed to a frozen TTS module that lies outside the training objective. This division keeps the learning problem centered on language modeling, rather than multimodal alignment or speech generation.

As illustrated in Figure~\ref{fig:architecture}, the interaction timeline is partitioned into discrete intervals called \emph{Units}. The Unit duration is a configurable hyperparameter; we use 1 second in our experiments (\S\ref{sec:exp}). Within each Unit, up to three channels operate concurrently:

\begin{itemize}[leftmargin=1.5em]
       \item \textbf{Listening (Perception) Channel}: Continuously ingests the user's audio stream, represented as a fixed number of audio tokens per second. This channel is \emph{always active}, enabling full-duplex interaction.
       \item \textbf{Visible Writing (Cognition) Channel}: \emph{Always active}, externalizing the model's text-native cognition as unconstrained free-form text. During listening it builds incremental reasoning that sharpens the subsequent response; during speaking it emits structured artifacts (code, tables, derivations) in parallel with natural speech.
       \item \textbf{Speaking (Expression) Channel}: Generates natural spoken language as a real-time oral response of the visible-writing output. This channel is active \emph{only during speaking}.
\end{itemize}

Each Unit exists in one of two states:
\begin{itemize}[leftmargin=1.5em]
       \item \textbf{Listening Unit}: Listening + Visible Writing (\texttt{ls\_cogn}). The model ingests audio and incrementally builds an understanding of the user's intent.
       \item \textbf{Speaking Unit}: Listening + Speaking + Visible Writing (\texttt{reply\_cogn}). The model simultaneously listens, speaks, and produces user-visible free-form text.
\end{itemize}

The three channels share a single attention context: all tokens (audio, visible-writing text, and spoken text) participate in the same causal attention computation. This implicit shared attention serves as the information fusion mechanism across channels, requiring no additional cross-channel alignment modules.

\subsection{Token schema design}
\label{sec:token_schema}

The core design principle of LWS is to encode the tri-channel visible-writing architecture entirely through \emph{token-level structure}, rather than architectural modifications. We define a set of special tokens that delineate channel boundaries within each Unit:

\paragraph{Listening Unit}
\begin{equation}
       \underbrace{
       \texttt{<unit>}\; \overbrace{a_1 \!\cdots\! a_{10}}^{\text{audio}}\;
       \texttt{<|lc|>}\; \overbrace{t_1 \!\cdots\! t_k}^{\text{writing}}\;
       \texttt{<|/lc|>}\; \texttt{</unit>}
       }_{\text{Listening Unit}}
\end{equation}

\paragraph{Speaking Unit}
\begin{equation}
       \underbrace{
       \texttt{<unit>}\; \overbrace{a_1 \!\cdots\! a_{10}}^{\text{audio}}\;
       \texttt{<|spk|>}\; \overbrace{s_1 \!\cdots\! s_m}^{\text{speak}}\;
       \texttt{<|eos|>}\; \texttt{<|rc|>}\; \overbrace{r_1 \!\cdots\! r_n}^{\text{writing}}\;
       \texttt{<|/rc|>}\; \texttt{</unit>}
       }_{\text{Speaking Unit}}
\end{equation}
We use abbreviated tags in equations for readability:

\vspace{-0.5em}
\begin{center}
       \small
       \begin{tabular}{@{}ll@{\qquad}ll@{}}
              \texttt{<|lc|>}  & \texttt{ls\_cogn\_bos} & \texttt{<|rc|>}  & \texttt{reply\_cogn\_bos} \\
              \texttt{<|/lc|>} & \texttt{ls\_cogn\_eos} & \texttt{<|/rc|>} & \texttt{reply\_cogn\_eos} \\
              \texttt{<|spk|>} & \texttt{speak}         & \texttt{<|eos|>} & \texttt{chunk\_eos}       \\
       \end{tabular}
\end{center}
\vspace{-0.5em}

\paragraph{How one Unit runs in practice}
During each interval, the model first receives the current audio chunk. If the interaction is still in the user's speaking phase, the Unit is instantiated as a Listening Unit: the model opens the \texttt{ls\_cogn} segment and emits incremental visible writing based only on the audio observed so far, such as partial understanding, disambiguation, or short notes. The user can therefore see text appear on screen even before the model begins speaking. Once the model enters the response phase, subsequent intervals become Speaking Units: the model continues ingesting the next audio chunk, emits spoken output in the \texttt{speak} segment, and produces richer visible writing in the \texttt{reply\_cogn} segment, such as code, tables, or structured analysis. Thus, across successive Units, LWS behaves as an always-listening system whose visible writing remains active throughout the interaction, while speech is activated only in those Units assigned to response generation.

\paragraph{Why separate the listening and speaking phases}
A natural alternative is to use a single unified cognition tag throughout the interaction. We instead separate the two cognitive phases because they are conditioned on different information states and occupy different positions in the causal timeline: one phase reflects ongoing interpretation of the user's incoming speech, while the other additionally conditions on the model's own emerging response. Making this distinction explicit simplifies next-token prediction and helps reduce information leakage across phases in full-duplex interaction. Empirically, without listen/reply separation the model did not learn stable timely response behavior, reaching 0.67 turn-taking TOR and 2.03 interruption score on 318 samples; a formal analysis is provided in Appendix~\ref{app:tag_analysis}.

\subsection{Inference}
\label{sec:inference}

Inference proceeds as a unit-level asynchronous full-duplex loop through an asynchronous pipeline. In each Unit, the incoming user audio chunk is encoded by a streaming Whisper encoder~\citep{radford2023robust} into audio embeddings, which are then fed to the LLM together with the structural delimiter \texttt{<unit>}. The first token generated by the LLM (\texttt{<|lc|>} or \texttt{<|spk|>}) determines whether the interval is realized as a Listening Unit or a Speaking Unit. In a Listening Unit, the model emits \texttt{ls\_cogn} text based only on the audio observed so far. In a Speaking Unit, the model first generates \texttt{speak} tokens; after the segment ends at \texttt{<|eos|>}, the corresponding hidden states are forwarded to an external TTS module, while the LLM continues generating \texttt{reply\_cogn} tokens in parallel. The Unit then closes and the next audio chunk begins. This design reduces response latency in three ways: audio is processed incrementally rather than after full-turn completion, speech synthesis starts as soon as a spoken segment is available, and visible writing can continue while TTS is already producing the spoken response. Because audio ingestion never stops during this process, user speech that arrives while the model is speaking is still encoded into subsequent Units, though it does not retroactively affect tokens already generated in the current Unit. This gives the system a built-in mechanism for interruption handling without relying on an external VAD. Consequently, the system sustains real-time interaction with minimal mode-switching overhead, while preserving continuity across listening, speaking, and visible-writing updates. Appendix~\ref{app:inference} provides the full pipeline diagram together with a complete unit-by-unit token trace.

\section{Data Construction}
\label{sec:data}
Training LWS requires data with per-second cognitive annotations aligned to an audio timeline, a format that does not exist in public corpora. We therefore design a two-stage pipeline (Figure~\ref{fig:data_construction}) that starts from standard text QA pairs and converts them into the Unit-based interaction format used by the model.

\textbf{Stage~1 (Offline cognitive synthesis)} begins from a text conversation pair and uses a strong teacher LLM to synthesize three text streams that can later be aligned to a second-by-second interaction timeline. The first stream, \texttt{streaming\_reasoning\_chain}, represents incremental understanding during the user's turn and is used to supervise \texttt{ls\_cogn}. Next, \texttt{voice\_response} provides a concise spoken-style paraphrase used to supervise \texttt{speak}. Finally, the original structured response is preserved as the supervision target for \texttt{reply\_cogn}, retaining exact written content. To make these streams consistent with streaming inference, we estimate a plausible speech duration, segment the interaction into per-second steps, and prompt the teacher with a strict causal constraint: at each second $t$, it may reason only from the portion of the input that would already have been heard at that point in the simulated timeline.

\textbf{Stage~2 (Online timeline construction)} combines the Stage~1 text streams with real audio recordings and their CTC-based character-level alignments to construct the final Unit sequence. We first build a global second-by-second timeline for each conversation, including random silence intervals between turns so that training examples reflect more natural conversational pacing. Each interval is then assigned as either a Listening Unit or a Speaking Unit according to who is currently speaking, and the corresponding text stream is inserted into the appropriate channel. Following the appendix construction, Listening Units are further categorized as \emph{early} (before user speech), \emph{effective} (during user speech), and \emph{post} (after user speech but before the model speaks), which lets the model learn when to wait, when to incrementally interpret the user, and when to prepare the reply. This distinction is important in full-duplex training because silence, active user speech, and post-turn transition periods require different visible-writing behavior even though all are represented as Listening Units. To expose the model to overlap and take-over behavior during training, we additionally apply interruption augmentation to a subset of conversations.

Full construction details and the Stage~1 prompt template are provided in Appendix~\ref{app:data_details} and~\ref{app:prompt}.

\section{Experiments}
\label{sec:exp}

\subsection{Experimental setup}

\paragraph{Base model}
We build LWS on top of the MiniCPM-V architecture~\citep{yao2024minicpm} with a Qwen3-8B backbone LLM~\citep{yang2025qwen3}. The Listening channel uses a streaming Whisper Medium encoder~\citep{radford2023robust} as the audio perception module (APM), with audio tokens produced at a rate of 10 tokens/second and pooled with a stride of 5. For the Speaking channel's TTS, we employ a LLaMA-based TTS backbone with FlashAttention~\citep{dao2022flashattention}, using S3Tokenizer~\citep{du2024cosyvoice} as the audio tokenizer. The TTS module conditions on both the LLM's output tokens and projected hidden states. The model is initialized from a duplex speech checkpoint and fine-tuned for the tri-channel LWS capability. In this final stage, the APM, TTS backbone, and hidden-state projection into TTS are frozen, and only the LLM parameters are updated.

\paragraph{Training data}
We construct training data using the two-stage pipeline described in \S\ref{sec:data}. The Stage~1 cognitive synthesis is applied to a joint interleaved multimodal corpus covering both Chinese and English.  After augmentation (including interruption augmentation), the final training set contains 500K examples. The data is formatted into the Unit-based token schema with 1-second Unit duration and interleaved audio and text chunks.

\paragraph{Training configuration}
We train on 32 NVIDIA A100 GPUs with bf16 mixed precision and gradient checkpointing enabled for both the full model and the LLM backbone. The LLM learning rate is $5 \times 10^{-6}$ with a cosine restart scheduler, minimum learning rate $1 \times 10^{-6}$, and linear warmup from $1 \times 10^{-8}$ over 200 steps. We use dynamic batching with a maximum sequence length of 4{,}096 tokens, gradient accumulation over 2 steps, and max\_steps=4{,}000. Only the LLM parameters are tuned; the APM and TTS module remain frozen throughout the final LWS training stage.

\paragraph{Unit duration} We set the Unit duration to 1 second (i.e., 1 fps) throughout all experiments. This choice balances writing output budget per Unit against real-time responsiveness. The architecture imposes no constraint on this value; shorter Units yield finer-grained interaction at the cost of less writing content per Unit, while longer Units allow deeper per-step reasoning but increase response granularity.

\subsection{Main results}
\subsubsection{Spoken understanding and reasoning}
\label{sec:uro_bench}

To evaluate spoken understanding and reasoning under multilingual and difficulty-stratified settings, we additionally report results on URO-Bench~\citep{yan2025uro}, whose Reasoning (R) dimension covers mathematical and logical reasoning tasks and thus directly measures structured reasoning ability. We compare LWS against representative public speech LLMs, and further include two separately trained internal ablations: \textit{w/o write while listen} (removing \texttt{ls\_cogn} supervision) and \textit{w/o write while speak} (removing \texttt{reply\_cogn} supervision). The public baseline numbers are taken from the Step-Audio 2 technical report~\citep{wu2025step} to keep the comparison aligned with the established benchmark protocol.

\begin{table*}[t]
       \centering
       
       \tablestyle
       \begin{tabular*}{\textwidth}{@{\extracolsep{\fill}}lcccccccc@{}}
              \toprule
              \multirow{2}{*}{\thdr{Method}} & \multicolumn{4}{c}{\thdr{Basic}} & \multicolumn{4}{c}{\thdr{Pro}}                                                                                              \\
              \cmidrule(lr){2-5} \cmidrule(lr){6-9}
              & \thdr{U}                         & \thdr{R}                       & \thdr{O}    & \thdr{Avg}  & \thdr{U}    & \thdr{R}    & \thdr{O} & \thdr{Avg}  \\
              \midrule
              \rowcolor{gray!12}
              \multicolumn{9}{@{}l}{\textbf{Chinese}}                                                                                                                                                \\
              GPT-4o-Audio                     & 89.4                               & 65.5                             & 85.2          & 78.6          & 70.6          & 57.2          & 70.2       & 67.1          \\
              GPT-Realtime                     & 88.8                               & 72.9                             & 90.8          & 80.6          & 72.3          & 62.6          & 74.2       & 70.6          \\
              Kimi-Audio                       & 79.3                               & 64.7                             & 79.8          & 73.6          & 60.4          & 59.3          & \textbf{76.2} & 66.0          \\
              Qwen-Omni                        & 59.7                               & 69.7                             & 77.3          & 69.0          & 59.0          & 59.8          & 58.7       & 59.1          \\
              Step-Audio 2                     & \textbf{91.1}                      & 75.5                             & 86.1          & \textbf{83.3} & 74.8          & 63.2          & 65.1       & 68.3          \\
              \midrule
              \rowcolor{gray!6}
              \textbf{Listen-Write-Speak}      & 79.3                               & 72.4                             & \textbf{96.1} & 82.6          & 92.5          & \textbf{85.9} & 75.5       & \textbf{84.6} \\
              w/o write while listen           & 65.9                               & 72.5                             & 95.3          & 77.9          & 86.0          & 84.4          & 74.6       & 81.7          \\
              w/o write while speak            & 73.4                               & \textbf{75.9}                    & 71.9          & 73.7          & \textbf{93.9} & 84.4          & 63.6       & 80.6          \\
              \midrule
              \rowcolor{gray!12}
              \multicolumn{9}{@{}l}{\textbf{English}}                                                                                                                                                \\
              GPT-4o-Audio                     & 90.2                               & 75.9                             & 90.4          & 84.5          & 60.7          & 64.4          & \textbf{78.5} & 67.5          \\
              GPT-Realtime                     & 87.4                               & \textbf{84.1}                    & \textbf{94.1} & \textbf{88.1} & 59.7          & 74.5          & 76.1       & 68.9          \\
              Kimi-Audio                       & 83.4                               & 42.3                             & 60.4          & 60.0          & 50.3          & 40.6          & 56.0       & 49.8          \\
              Qwen-Omni                        & 66.3                               & 69.6                             & 76.2          & 70.6          & 44.5          & 63.9          & 49.4       & 51.0          \\
              Step-Audio 2                     & \textbf{92.7}                      & 76.5                             & 84.9          & 83.9          & 64.9          & 67.8          & 66.3       & 66.1          \\
              \midrule
              \rowcolor{gray!6}
              \textbf{Listen-Write-Speak}      & 83.3                               & 69.6                             & 92.7          & 81.9          & 74.6          & \textbf{89.0} & 70.3       & \textbf{78.0} \\
              w/o write while listen           & 76.6                               & 72.2                             & 92.4          & 80.4          & \textbf{75.4} & 84.6          & 66.8       & 75.6          \\
              w/o write while speak            & 74.8                               & 72.2                             & 66.3          & 71.1          & 73.0          & 81.3          & 52.2       & 68.8          \\
              \bottomrule
       \end{tabular*}
\caption{URO-Bench results comparing representative speech LLM baselines and our three variants. \textbf{U}: understanding, \textbf{R}: reasoning, \textbf{O}: oral. Higher is better. Best score within each language block and column is in bold.}
       \label{tab:uro_bench}
\end{table*}

Table~\ref{tab:uro_bench} shows a mixed but informative pattern. Compared with public baselines, LWS is especially strong on harder Chinese splits, achieving the best Chinese Pro average (84.6), the best Chinese Pro understanding and reasoning scores (92.5/85.9), and the highest Chinese oral-basic score (96.1). On English Basic, GPT-Realtime remains strongest in average score (88.1), but LWS remains competitive. Within our model family, LWS outperforms both ablations on all four displayed Basic/Pro averages across Chinese and English (Chinese Basic: 82.6 vs.\ 77.9/73.7; Chinese Pro: 84.6 vs.\ 81.7/80.6; English Basic: 81.9 vs.\ 80.4/71.1; English Pro: 78.0 vs.\ 75.6/68.8), supporting the value of visible writing during both listening and speaking phases in this benchmark setting. Table~\ref{tab:reasoning_ablation} further summarizes the reason-chain/writing versus speak-only training comparison on Spoken-MQA and aggregate URO-Bench metrics reported below.

\begin{table}[t]
       \centering
       
       {\footnotesize
       \renewcommand{\arraystretch}{1.08}
       \setlength{\tabcolsep}{4pt}
       \begin{tabular*}{\columnwidth}{@{\extracolsep{\fill}}lccc@{}}
              \toprule
              \textbf{Metric} & \textbf{Writing} & \textbf{Speak-only} & \textbf{Gain} \\
              \midrule
              \rowcolor{gray!12}
              \multicolumn{4}{@{}l}{\textbf{Spoken-MQA}} \\
              long digit & \textbf{39.31} & 10.98 & \textbf{+28.33} \\
              short digit & \textbf{37.00} & 24.00 & \textbf{+13.00} \\
              single-step reasoning & \textbf{63.97} & 59.26 & \textbf{+4.71} \\
              multi-step reasoning & \textbf{50.07} & 40.80 & \textbf{+9.27} \\
              \rowcolor{gray!6}
              \textbf{Overall} & \textbf{52.31} & 42.62 & \textbf{+9.69} \\
              \midrule
              \rowcolor{gray!12}
              \multicolumn{4}{@{}l}{\textbf{URO-Bench aggregate}} \\
              Overall Macro & \textbf{80.7} & 70.0 & \textbf{+10.7} \\
              BASIC Average & \textbf{80.7} & 72.3 & \textbf{+8.4} \\
              PRO Average & \textbf{79.5} & 66.4 & \textbf{+13.1} \\
              \bottomrule
       \end{tabular*}}

\caption{Reason-chain / visible-writing training versus a speak-only training baseline on spoken reasoning and aggregate URO-Bench metrics. Higher is better; best scores are bolded within each comparison block.}
       \label{tab:reasoning_ablation}
\end{table}

\begin{table}[t]
       \centering
       
       {\footnotesize
       \renewcommand{\arraystretch}{1.06}
       \setlength{\tabcolsep}{3pt}
       \begin{tabular*}{\columnwidth}{@{\extracolsep{\fill}}lccc@{}}
              \toprule
              \textbf{Dataset} & \textbf{Writing} & \textbf{Speak-only} & \textbf{Gain} \\
              \midrule
              \rowcolor{gray!12}
              \multicolumn{4}{@{}l}{\textbf{Math / reasoning and structured interaction}} \\
              Gsm8kEval & \textbf{76.8} & 61.3 & \textbf{+15.5} \\
              GaokaoEval & \textbf{94.1} & 80.5 & \textbf{+13.6} \\
              CodeSwitching-en & \textbf{82.0} & 52.9 & \textbf{+29.1} \\
              CodeSwitching-zh & \textbf{92.6} & 68.3 & \textbf{+24.3} \\
              OpenbookQA-zh & \textbf{84.7} & 72.5 & \textbf{+12.2} \\
              \midrule
              \rowcolor{gray!12}
              \multicolumn{4}{@{}l}{\textbf{Open-ended and safety-oriented tasks}} \\
              Summary & \textbf{92.7} & 84.6 & \textbf{+8.1} \\
              Safety-en & \textbf{96.8} & 81.6 & \textbf{+15.2} \\
              Safety-zh & \textbf{96.8} & 80.8 & \textbf{+16.0} \\
              StoralEval & \textbf{79.2} & 57.9 & \textbf{+21.3} \\
              WildchatEval & \textbf{94.7} & 71.2 & \textbf{+23.5} \\
              \bottomrule
       \end{tabular*}}
\caption{Representative URO-Bench dataset results. The writing model is strongest on math, code-switching, safety, and open-ended conversational subsets. Higher is better; best scores are bolded.}
       \label{tab:uro_dataset_ablation}
\end{table}

\subsubsection{Reply quality}
\label{sec:reply_quality}

We further assess reply quality on VoiceBench~\citep{chen2024voicebench}, reported in Table~\ref{tab:voicebench_comparison}. VoiceBench follows a speech-to-text evaluation protocol: the model receives spoken input, but the scored output is \emph{text}. The main LWS row uses \texttt{reply\_cogn + speak} as the judged prediction for non-AlpacaEval metrics. We also report a \textit{speak-only} row using only the spoken field from the same model; this is an output-field comparison rather than an inference-time channel ablation. LWS reaches 4.72 on AlpacaEval and 73.79 overall on full VoiceBench.

\begin{table*}[t]
       \centering
       
       {\scriptsize
       \renewcommand{\arraystretch}{1.1}
       \setlength{\tabcolsep}{2.3pt}
       \begin{tabular*}{\textwidth}{@{\extracolsep{\fill}}lcccccccccc@{}}
              \toprule
              \textbf{Model} & \textbf{Alpaca} & \textbf{Common} & \textbf{Wild} & \textbf{SD-QA} & \textbf{MMSU} & \textbf{OBQA} & \textbf{BBH} & \textbf{IFEval} & \textbf{Adv.} & \textbf{Overall} \\
              \midrule
              \rowcolor{gray!6}
              \textbf{LWS (reply+speak)} & \textbf{4.72} & \textbf{4.10} & \textbf{4.22} & 52.41 & \textbf{67.86} & 82.42 & \textbf{64.00} & \textbf{37.39} & 99.23 & \textbf{73.79} \\
              LWS (speak only) & 4.21 & 3.53 & 3.40 & 48.23 & 67.01 & \textbf{82.64} & 57.30 & 19.13 & 94.23 & 65.70 \\
              \midrule
              \rowcolor{gray!12}
              \multicolumn{11}{@{}l}{\textbf{Full-duplex / open speech baselines}} \\
              Nemotron 3 VoiceChat & 3.59 & 3.12 & 3.06 & 43.40 & 50.46 & 64.40 & 52.70 & 17.12 & \textbf{99.62} & 58.10 \\
              Freeze-Omni & 4.03 & 3.46 & 3.15 & \textbf{53.45} & 28.14 & 30.98 & 50.70 & 23.40 & 97.30 & 55.20 \\
              Moshi & 2.01 & 1.60 & 1.30 & 15.64 & 24.04 & 25.93 & 47.40 & 10.12 & 44.23 & 29.51 \\
              \bottomrule
       \end{tabular*}}
\caption{Full VoiceBench results. AlpacaEval, CommonEval, and WildVoice use the original 1 to 5 OpenQA scale; other metrics and Overall follow the VoiceBench percentage convention. LWS reply+speak uses \texttt{reply\_cogn + speak}; LWS speak-only uses only the spoken field from the same model. Best scores are bolded.}
       \label{tab:voicebench_comparison}
\end{table*}
\subsubsection{Channel consistency}
\label{sec:channel_consistency}

A potential risk of producing both \texttt{speak} and \texttt{reply\_cogn} from the same context is \emph{self-contradiction}: the spoken response may diverge from the visible written response. We evaluate this risk on AlpacaEval~\citep{alpaca_eval} samples from VoiceBench~\citep{chen2024voicebench} using an LLM-as-judge protocol: GPT-5 receives the user query, Visible Writing output (\texttt{reply\_cogn}), and Speaking output (\texttt{speak}), and assigns \textbf{1} if the spoken answer is factually consistent with the written one and \textbf{0} otherwise. LWS achieves a \textbf{channel consistency rate of 92.6\%} (589/636), indicating that the two user-facing channels remain well aligned in the large majority of cases.

\subsubsection{Full-duplex interaction}

\begin{table*}[t]
       \centering
       
       \scriptsize
       \renewcommand{\arraystretch}{1.1}
       \setlength{\tabcolsep}{3pt}
       \begin{tabular*}{\textwidth}{@{\extracolsep{\fill}}p{2.3cm}cccccccccc@{}}
              \toprule
              \multirow{2}{*}{\thdr{Model}} & \multicolumn{2}{c}{\thdr{Pause}} & \multicolumn{3}{c}{\thdr{Backchannel}} & \multicolumn{2}{c}{\thdr{Turn Taking}} & \multicolumn{3}{c}{\thdr{Interruption}} \\
              \cmidrule(lr){2-3} \cmidrule(lr){4-6} \cmidrule(lr){7-8} \cmidrule(lr){9-11}
              & \theadcell{Syn.TOR$\downarrow$}
              & \theadcell{Can.TOR$\downarrow$}
              & \theadcell{TOR$\downarrow$}
              & \theadcell{Freq$\uparrow$}
              & \theadcell{JSD$\downarrow$}
              & \theadcell{Can.TOR$\uparrow$}
              & \theadcell{Lat.$\downarrow$}
              & \theadcell{TOR$\uparrow$}
              & \theadcell{Score$\uparrow$}
              & \theadcell{Lat.$\downarrow$} \\
              \midrule
              \rowcolor{gray!12}
              \multicolumn{11}{@{}l}{\textbf{Public baselines}} \\
              PersonaPlex                   & 0.58 & 0.66 & 0.33 & \textbf{0.03} & \textbf{0.65} & 0.99 & \textbf{0.07} & \textbf{1.00} & \textbf{4.21} & 0.40 \\
              dGSLM                         & 0.93 & 0.94 & 0.69 & 0.02 & 0.93 & 0.98 & 0.35 & 0.92 & 0.20 & 2.53 \\
              Moshi                         & 0.98 & 0.98 & 1.00 & 0.00 & 0.96 & 0.94 & 0.27 & \textbf{1.00} & 0.77 & \textbf{0.26} \\
              Freeze-Omni                   & 0.64 & 0.48 & 0.64 & 0.00 & 1.00 & 0.34 & 0.95 & 0.87 & 3.62 & 1.41 \\
              \makecell[l]{Gemini Live 2.0\\(deprecated)} & 0.26 & 0.31 & 0.09 & 0.01 & 0.90 & 0.66 & 1.30 & 0.89 & 3.38 & 1.18 \\
              GPT-Realtime                  & \textbf{0.01} & 0.12 & \textbf{0.00} & 0.01 & 0.98 & \textbf{1.00} & 1.47 & 0.97 & 3.85 & 1.50 \\
              \cmidrule(lr){1-11}
              \textbf{LWS}                  & \textbf{0.01} & \textbf{0.01} & 0.53 & 0.00 & 1.00 & 0.97 & 0.48 & 0.99 & 4.02 & 0.65 \\
              \bottomrule
       \end{tabular*}
\caption{Full-Duplex-Bench results across four interaction settings. \textbf{TOR}: take-over rate. \textbf{FRQ}: backchannel frequency. \textbf{JSD}: Jensen-Shannon divergence. Arrows indicate whether lower or higher is better.}
       \label{tab:full_duplex_bench}
\end{table*}

Table~\ref{tab:full_duplex_bench} summarizes Full-Duplex-Bench results. In smooth turn-taking, Listen-Write-Speak reaches 0.97 Candor TOR at 0.48\,s latency, slightly behind the fastest specialized models such as PersonaPlex and Moshi, but substantially faster than larger commercial realtime models while maintaining competitive turn-taking quality. Under user interruption, it attains 4.02 GPT-4o quality at 0.65\,s latency, indicating that it can recover from overlap and continue with coherent responses without incurring the large delay observed in several baselines. On pause handling, it also matches the best reported TOR (0.01) on both the synthetic and Candor settings. This pattern is consistent with the design goal of keeping \texttt{ls\_cogn} active during listening: the model can continuously form a response plan before taking the turn, which helps preserve response quality during overlapping, rapidly changing interactions in practice.

\subsubsection{Training convergence}

\begin{figure}[t]
       \centering
       \includegraphics[width=\textwidth]{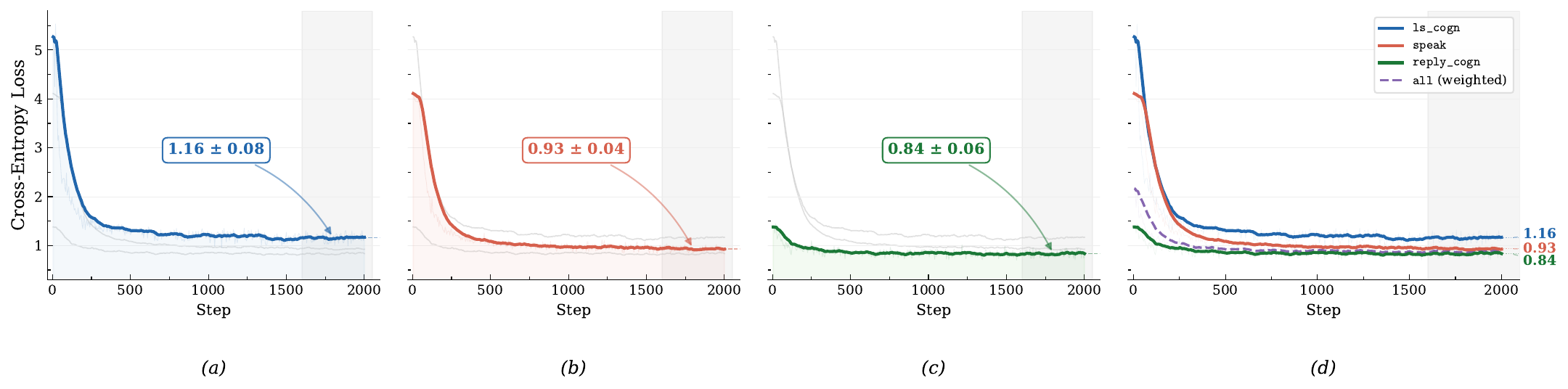}
       \caption{Per-channel training loss curves for \texttt{ls\_cogn}, \texttt{speak}, \texttt{reply\_cogn}, and their weighted overlay, showing smooth convergence without optimization instability.}
       \label{fig:training_loss}
\end{figure}

A fundamental question is whether the tri-channel token schema can be stably trained within a single autoregressive LLM. As shown in Figure~\ref{fig:training_loss}, we decompose training loss by channel. All three channels converge smoothly with low variance (coefficient of variation $< 8\%$). The converged losses are 1.16 (\texttt{ls\_cogn}), 0.93 (\texttt{speak}), and 0.84 (\texttt{reply\_cogn}), respectively. \texttt{reply\_cogn} converges fastest and to the lowest loss, consistent with its role as standard text generation conditioned on rich context. \texttt{ls\_cogn} starts highest but converges stably, reflecting the greater difficulty of incremental audio comprehension. The overlay in Figure~\ref{fig:training_loss}(d) shows no obvious optimization instability in the joint setting, supporting the feasibility of tri-channel joint training within a single autoregressive LLM in our setup.

\section{Limitations}
\label{sec:limitations}

LWS combines full-duplex listening, visible writing, and real-time speaking within a standard autoregressive framework, but the present study remains subject to several limitations. (1)~\textbf{Reasoning depth remains constrained by real-time operation.} Because visible writing and speech must be produced within each Unit, the system is optimized for responsiveness and is not yet well suited to longer-horizon reasoning, multi-step planning, or complex tool-mediated workflows, where stronger performance may require slower response generation or an explicit mechanism for deferring speech while deeper written reasoning unfolds. (2)~\textbf{The input interface remains narrow.} The current formulation assumes speech-only user input, limiting applicability in settings where users naturally combine speech with code, screenshots, tables, or other context. Extending LWS to adaptive multimodal input is therefore an important direction for future work.

\section{Conclusion}
\label{sec:conclusion}

We presented Listen-Write-Speak (LWS), a text-first tri-channel paradigm that unifies full-duplex listening, visible writing, and real-time speaking within a single autoregressive LLM. The design is realized entirely through a token schema without architectural changes, supported by a practical two-stage data pipeline. Empirically, LWS performs strongly on Full-Duplex-Bench, reaches 4.72 on VoiceBench AlpacaEval, achieves 92.6\% writing and speaking consistency, and outperforms its internal ablations on multilingual URO-Bench.

\section*{Ethics Statement}
Listen-Write-Speak changes not only \emph{what} a speech model outputs, but also \emph{how} users may trust it: because LWS can respond fluently in speech while simultaneously producing polished written artifacts (e.g., code, structured analysis, or derivations), it may increase the perceived authority and actionability of incorrect, biased, or unsafe content, especially in real-time settings where users may act on the spoken summary before fully inspecting the written output. The visible-writing channel improves inspectability relative to speech-only models, but it is not itself a guarantee of correctness, faithfulness, or safety; accordingly, we recommend deployment with synchronized moderation across both spoken and written channels, policy-based refusal for hazardous requests, logging and auditing of model outputs, and clear user disclosure that visible writing is an assistive intermediate artifact rather than verified ground truth. Our training pipeline uses synthetic per-second annotations constructed from existing text and audio resources and does not rely on private user conversations, but any real-world deployment that processes live voice should still be governed by consent, retention, and access-control policies, since spoken interactions may contain sensitive personal information. We used external LLMs only as data-construction and evaluation tools: Qwen3 235B was used in Stage~1 to synthesize per-second cognitive annotations and spoken-style paraphrases under the causal prompt template in Appendix~\ref{app:prompt}, and GPT-5 was used as an LLM judge for the channel-consistency analysis in \S\ref{sec:channel_consistency}; the reply-quality evaluation prompts are listed in Appendix~\ref{app:eval_prompts}. These external LLMs are not components of LWS itself. We aim to release enough detail to support reproducibility while avoiding operational details that would materially lower the barrier to misuse.


\bibliographystyle{colm2026_conference}
\bibliography{colm2026_conference}

\appendix
\section{Inference Pipeline and Token Trace}
\label{app:inference}

\begin{figure}[h]
       \centering
       \includegraphics[width=\textwidth]{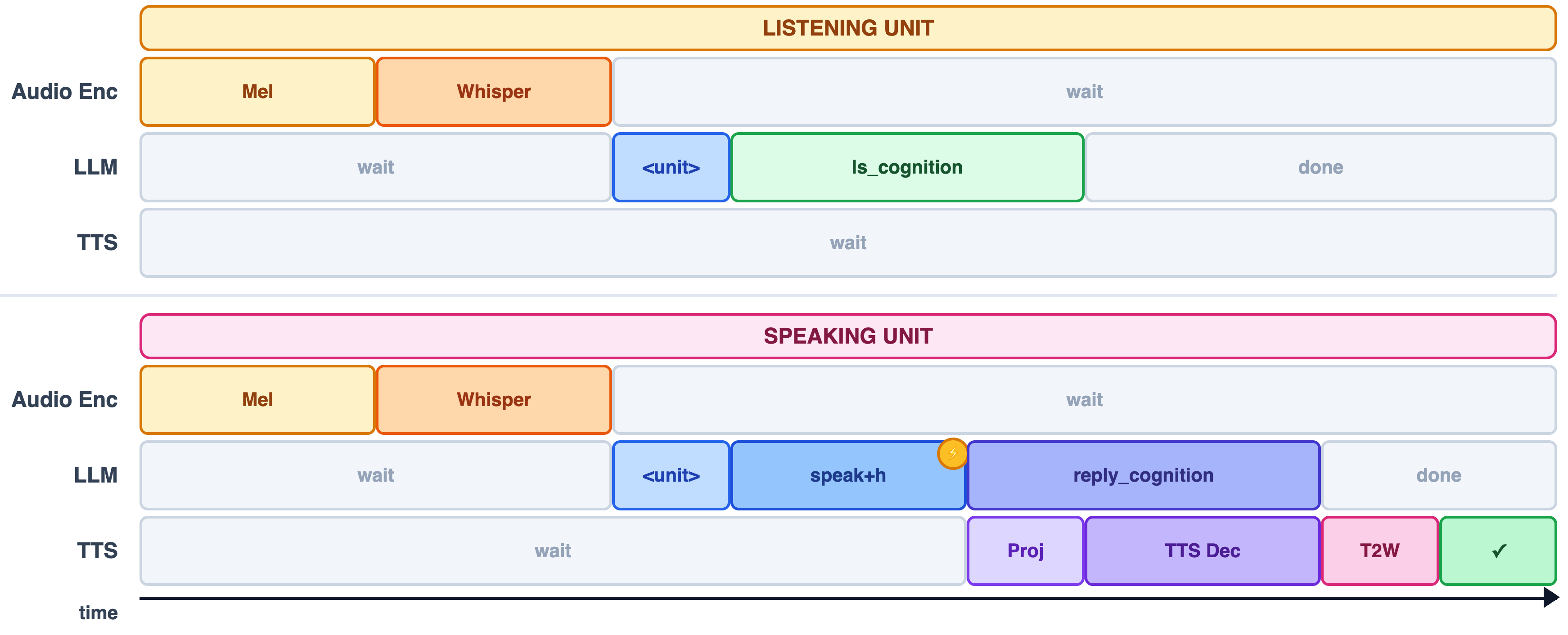}
       \caption{LWS inference pipeline. Three asynchronous threads process each Unit: the LLM generates speaking and visible-writing tokens, TTS synthesizes speech from speaking hidden states, and the audio encoder continuously streams user input.}
       \label{fig:Inference}
\end{figure}

Figure~\ref{fig:Inference} illustrates the full inference pipeline. In each Unit, the user's audio is encoded by a streaming Whisper encoder~\citep{radford2023robust} into audio embeddings, which are fed to the LLM together with a \texttt{<unit>} delimiter. The LLM then autoregressively generates the Unit's content: the first token it produces (\texttt{<|lc|>} or \texttt{<|spk|>}) autonomously determines the Unit type (Listening or Speaking). In a Speaking Unit, speaking tokens are generated first; once the speaking segment ends at \texttt{<|eos|>}, its hidden states are immediately forwarded to an external TTS module running in a separate thread, and the LLM continues generating visible-writing tokens (\texttt{reply\_cogn}) \emph{in parallel} with TTS synthesis. The Unit closes with \texttt{</unit>} and the next Unit begins.

\paragraph{Complete token sequence example.}
We present a unit-by-unit token trace for a single-turn interaction. The user asks via speech for a summary of \emph{One Hundred Years of Solitude} by Gabriel Garc\'ia M\'arquez. The trace uses the abbreviated notation from \S\ref{sec:token_schema}.

\noindent\texttt{<|im\_start|>system}\\
\texttt{You are a Listen-Write-Speak model with three channels: listening, visible writing, and speaking.}\\
\texttt{<|im\_end|>}\\
\texttt{<|im\_start|>user}\\
\texttt{<|audio\_start|>}$\;a_1 \cdots a_N\;$\texttt{<|audio\_end|><|im\_end|>}

\paragraph{Listening phase (Units 1 to 6).}
Each Listening Unit ingests audio tokens and generates \texttt{ls\_cogn} output. The per-Unit structure is: \texttt{<unit>}\,$a_1 \!\cdots\! a_{10}$\,\texttt{<|lc|>}\,\emph{content}\,\texttt{<|/lc|>}\,\texttt{</unit>}. Units~1 to 2 produce empty visible-writing output while audio is still arriving; from Unit~3 onward, the model incrementally interprets the user's intent:

\begin{table}[h]
       \centering
       
       \tablestyle
       \begin{tabular}{@{}cp{0.72\columnwidth}@{}}
              \toprule
              \textbf{Unit} & \textbf{\texttt{ls\_cogn} content}                                                       \\
              \midrule
              1             & \emph{(empty)}                                                                           \\
              2             & \emph{(empty)}                                                                           \\
              3             & The user is asking for a short piece on Gabriel Garc\'ia M\'arquez's novel.              \\
              4             & The novel's full title is being revealed: `One Hundred Years of Solitude'. The           \\
              5             & full title is now complete, confirming the subject is the novel by Garc\'ia              \\
              6             & M\'arquez. The user wants a concise, insightful summary of the novel's themes and style. \\
              \bottomrule
       \end{tabular}
\caption{Listening phase: per-Unit \texttt{ls\_cogn} content.}
       \label{tab:listen_trace}
\end{table}

\paragraph{Speaking phase (Units 7 to 19).}
Upon detecting end-of-turn, the model transitions to Speaking Units. Each Unit simultaneously generates \texttt{speak} tokens (streamed to TTS for real-time voice) and \texttt{reply\_cogn} tokens (displayed as on-screen text). Table~\ref{tab:speak_trace} shows the parallel channel outputs.

\begin{table*}[h]
       \centering
       
       \tablestyle
       \begin{tabular}{@{}cll@{}}
              \toprule
              \textbf{Unit} & \textbf{Speaking (\texttt{speak})}  & \textbf{Visible Writing (\texttt{reply\_cogn})}   \\
              \midrule
              7             & Sure! *One Hundred                  & *One Hundred Years of Solitude*                   \\
              8             & Years of Solitude* is               & by Gabriel Garc\'ia M\'arquez is a                \\
              9             & a magical real                      & landmark of magical realism, chronicling          \\
              10            & ist masterpiece about the           & seven generations of the Buend\'ia family         \\
              11            & Buend\'ia family                    & in the mythical town of Macondo.                  \\
              12            & in Macondo,                         & Blending myth, history, and fantasy,              \\
              13            & blending myth and                   & the novel explores themes of time,                \\
              14            & reality to explore                  & fate, and isolation, weaving a                    \\
              15            & time,                               & dreamlike narrative where the extraordinary feels \\
              16            & fate, and isolation                 & ordinary. With lyrical prose and rich             \\
              17            & through a                           & symbolism, Garc\'ia M\'arquez creates an          \\
              18            & dreamlike, ly                       & enduring tale of love, madness, and               \\
              19            & rical story. \texttt{<|turn\_eos|>} & the inescapable cycles of human experience.       \\
              \bottomrule
       \end{tabular}
\caption{Speaking phase: per-Unit parallel channel outputs. The Speaking channel produces a concise oral response while the Visible Writing channel simultaneously generates a richer literary analysis.}
       \label{tab:speak_trace}
\end{table*}

\paragraph{Decoded outputs.}
Concatenating each channel's per-Unit tokens yields the final user-facing content:

\begin{itemize}[leftmargin=1.5em]
       \item \textbf{Speaking (\texttt{speak}):} ``Sure! One Hundred Years of Solitude is a magical realist masterpiece about the Buend\'ia family in Macondo, blending myth and reality to explore time, fate, and isolation through a dreamlike, lyrical story.''
       \item \textbf{Visible Writing (\texttt{reply\_cogn}):} ``One Hundred Years of Solitude by Gabriel Garc\'ia M\'arquez is a landmark of magical realism, chronicling seven generations of the Buend\'ia family in the mythical town of Macondo. Blending myth, history, and fantasy, the novel explores themes of time, fate, and isolation, weaving a dreamlike narrative where the extraordinary feels ordinary. With lyrical prose and rich symbolism, Garc\'ia M\'arquez creates an enduring tale of love, madness, and the inescapable cycles of human experience.''
\end{itemize}

\noindent The asymmetry is evident: \texttt{speak} produces a one-sentence response suitable for spoken delivery, while \texttt{reply\_cogn} simultaneously generates a detailed multi-sentence literary analysis displayed as on-screen text. This illustrates the core principle that visible writing is the primary, unconstrained output modality.

\section{Data Construction Details}
\label{app:data_details}

This appendix provides the full details of the two-stage data construction pipeline summarized in \S\ref{sec:data}.

\begin{figure}[H]
       \centering
       \includegraphics[width=\textwidth]{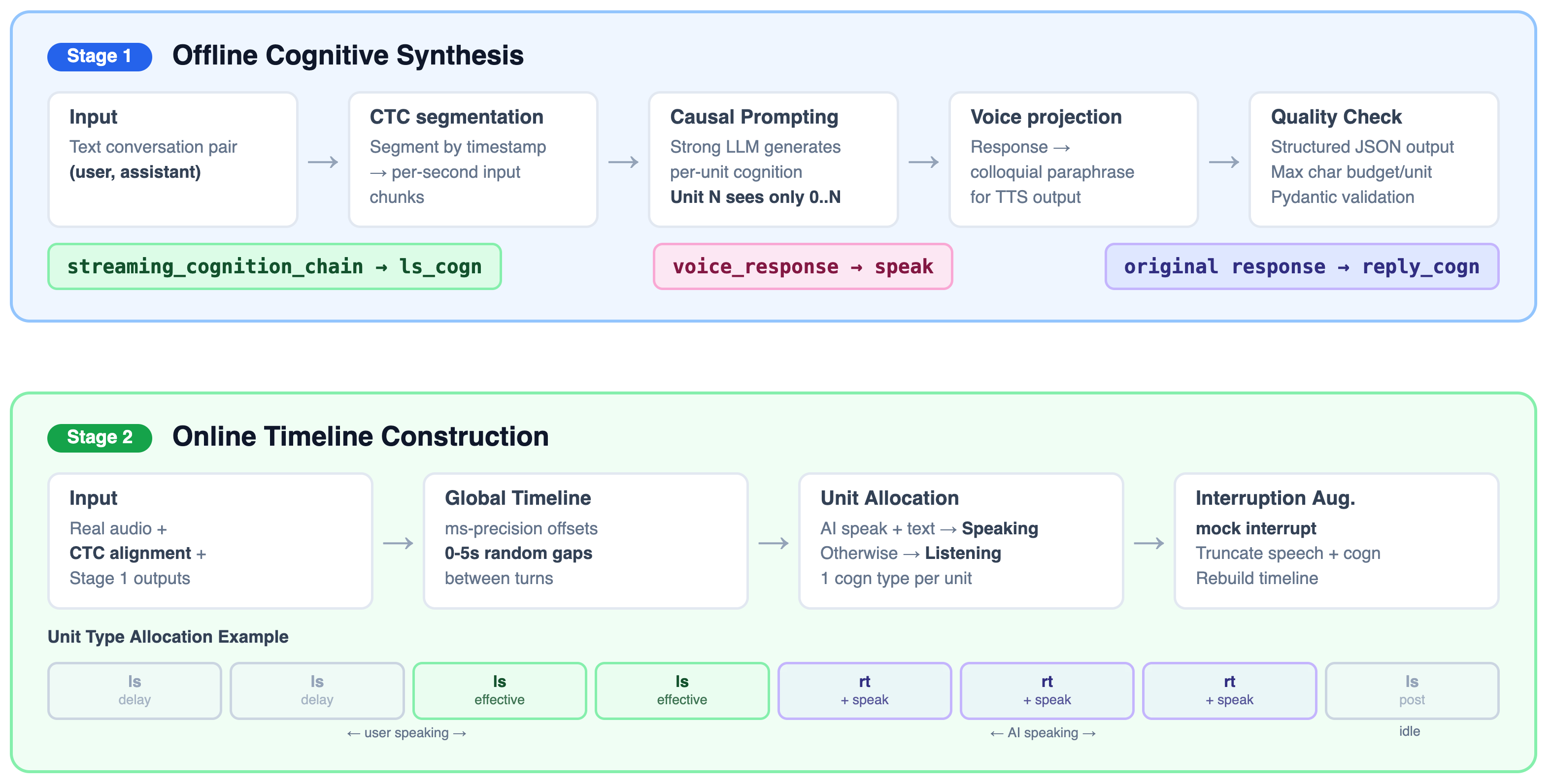}
       \caption{Two-stage data construction pipeline. Stage~1 uses a teacher LLM to synthesize per-second cognitive annotations from text QA pairs. Stage~2 aligns these annotations with real audio via CTC timestamps and constructs the final Unit-based token sequences, with optional interruption augmentation.}
       \label{fig:data_construction}
\end{figure}

\subsection{Stage 1: Offline cognitive synthesis}
\label{app:stage1}

Given a text conversation pair, we use a strong text LLM to generate per-second cognitive annotations intended to be causally consistent with the revealed input timeline.

\paragraph{Speech rate estimation.} We first estimate the number of seconds the spoken response will occupy, using empirical speech rates: ${\sim}7$ characters/second for Chinese and ${\sim}3.5$ words/second for English. The response is segmented into per-second chunks accordingly.

\paragraph{Causal constraint prompting.} We prompt the LLM to simulate a \emph{streaming} reasoning process: at each second $t$, the model generates writing text based only on what has been ``heard'' up to second $t$ (for listening-phase writing, \texttt{ls\_cogn}) or what has been ``said'' up to second $t$ (for reply-phase writing, \texttt{reply\_cogn}). This prompt enforces a causal constraint that prevents the model from using future information.

\paragraph{Quality control.} We use structured output (Pydantic schema) to guarantee format validity, and filter samples where per-second writing text exceeds a character budget or the voice response is empty.

\subsection{Stage 2: Online timeline construction}
\label{app:stage2}

Stage~2 takes real audio recordings and CTC-based character-level alignments, together with the Stage~1 text outputs, and constructs the final token sequences.

\paragraph{Global timeline construction.} We build a second-by-second timeline for the entire multi-turn conversation, inserting random silence intervals between turns to simulate natural conversational rhythm.

\paragraph{Unit type allocation.} Each Unit is assigned a type based on who is speaking and whether speech text is available:
\begin{itemize}[leftmargin=1.5em]
       \item If the AI is speaking and has \texttt{speak} text $\rightarrow$ \textbf{Speaking Unit} (with \texttt{reply\_cogn})
       \item Otherwise $\rightarrow$ \textbf{Listening Unit} (with \texttt{ls\_cogn})
\end{itemize}
Listening Units are further categorized as \emph{early} (user not yet speaking), \emph{effective} (user speaking), and \emph{post} (after user finishes, before AI speaks), each with appropriate writing content.

\section{Why separate the listening and speaking phases?}
\label{app:tag_analysis}

We provide the formal justification for using separate tags for listening-phase writing (\texttt{ls\_cogn}) and reply-phase writing (\texttt{reply\_cogn}), rather than a single unified \texttt{<cogn>} tag.

\paragraph{Different conditional distributions.}
The two writing modes are conditioned on fundamentally different information:
\begin{itemize}[leftmargin=1.5em]
       \item \texttt{ls\_cogn} (listening-phase writing): $P_{\text{lc}}(y_t \mid \mathbf{a}_{\leq T}, \mathbf{h})$, conditioned only on audio history and prior writing. Its role is incremental understanding of the user's intent.
       \item \texttt{reply\_cogn} (reply-phase writing): $P_{\text{rc}}(y_t \mid \mathbf{a}_{\leq T}, \mathbf{s}_{\leq T}, \mathbf{h})$, additionally conditioned on the model's own speech output. Its role is producing structured output (code, analysis, Markdown) that goes beyond what is being said aloud.
\end{itemize}

\paragraph{Explicit state reduces conditional entropy.}
Let $s_t \in \mathcal{S} = \{\text{lc},\, \text{rc},\, \text{spk}\}$ denote the semantic state of token $y_t$, corresponding to listening-phase writing, reply-phase writing, and speaking respectively. With a unified tag, the model must implicitly infer the state, facing a mixture distribution:
\begin{equation}
       P_{\text{mixed}}(y_t) = \sum_{s \in \mathcal{S}} \pi_s \cdot P_s(y_t)
\end{equation}
where $\pi_s = P(s \mid y_{<t}, \mathbf{x})$. By Jensen's inequality applied to entropy's concavity:
\begin{equation}
       H(P_{\text{mixed}}) \geq \sum_{s} \pi_s \cdot H(P_s)
\end{equation}
Since the three state-conditional distributions differ substantially (\texttt{ls\_cogn} produces incremental understanding, \texttt{reply\_cogn} produces code or structured analysis, and \texttt{speak} produces natural spoken language), the KL divergences $D_{\text{KL}}(P_s \| P_{s'})$ are large, and the inequality is strict. Explicit state markers eliminate this mixture penalty, reducing the per-token prediction entropy.

\paragraph{Causal consistency in full-duplex.}
Full-duplex interaction imposes a strict temporal causal constraint: the model's speech at time $t$ may only depend on user audio up to time $t$ and prior writing, never on future input. The token schema encodes this causal chain explicitly:
\begin{equation}
       \texttt{ls\_cogn} \;\rightarrow\; \texttt{reply\_cogn} \;\rightarrow\; \texttt{speak}
\end{equation}
The \texttt{<|/lc|>} token marks the boundary where listening-phase writing ends and reply-phase writing begins, ensuring that speaking generation cannot leak information from a listening-phase writing segment that has not yet concluded. With a unified \texttt{<cogn>} tag, the boundary between ``still understanding the user'' and ``already producing the reply'' is ambiguous, creating a risk of causal violation during training.

\section{Data Construction Prompt Template}
\label{app:prompt}

We provide the complete prompt template used in Stage~1 (\S\ref{app:stage1}) to synthesize per-second causal cognitive annotations from text QA pairs using Qwen3 235B. The prompt enforces a strict causal constraint: at each second $t$, the model may only reason about input received up to second $t$.

\paragraph{Input preparation.} Given a user instruction, we first segment it into per-second chunks using speech-rate estimation (7 Chinese characters/s or 3.5 English words/s). Let $T$ denote the total number of seconds. Substantive reasoning begins at second $t_0 = \min(2, T)$; earlier seconds serve as ASR-style confirmation.

\begin{promptbox}[System Prompt]
       \small\ttfamily
       You are an AI assistant that emulates streaming reasoning.\\[4pt]
       \textnormal{\textbf{\textsf{\#\# Context}}}\\
       User audio arrives second by second. At second N you only know text from second 0 to N.\\[4pt]
       \textnormal{\textbf{\textsf{\#\# Output Structure}}}\\
       1. \textasciigrave streaming\_reasoning\_chain\textasciigrave\ (\{T - t\textsubscript{0}\} entries):\\
       \quad - \textasciigrave second\textasciigrave: \{t\textsubscript{0}\} \textasciitilde\ \{T - 1\}\\
       \quad - \textasciigrave reasoning\_this\_second\textasciigrave: based only on known input, \{C-10\}\textasciitilde\{C+10\} characters long\\
       2. \textasciigrave voice\_response\textasciigrave: a concise, friendly spoken reply based on the complete text response, length should be around 30\%\textasciitilde 40\% of the original response. Avoid introducing details that are not in the original response.\\[4pt]
       \textnormal{\textbf{\textsf{\#\# Instructions}}}\\
       - Keep per-second reasoning length stable; even early seconds should stay in range.\\
       - Early seconds may be ASR-style confirmations ("I heard...") while waiting for more input.\\
       - Never mention content that arrives after the current second.\\
       - streaming\_reasoning\_chain should start producing substantive reasoning at second \{t\textsubscript{0}\}.\\
       - The final voice response must reflect the full written reply.
\end{promptbox}

\noindent where $T$ is the total number of user-input seconds, $t_0$ is the reasoning start second, and $C$ is the target characters per second (default 40).

\begin{promptbox}[User Prompt]
       \small\ttfamily
       \#\#\# User input by second:\\
       \quad Second 0: "\{segment\_0\}"\\
       \quad Second 1: "\{segment\_1\}"\\
       \quad ...\\
       \quad Second \{T-1\}: "\{segment\_\{T-1\}\}"\\[4pt]
       \#\#\# Written reply:\\
       \{original\_response\}\\[4pt]
       Follow the instructions and emit \textasciigrave StreamingColloquialRewrite\textasciigrave.\\
       Streaming reasoning starts at second \{t\textsubscript{0}\}.
\end{promptbox}

\noindent The model is required to return a structured JSON object (enforced via Pydantic schema \texttt{StreamingColloquialRewrite}) containing: (1)~\texttt{streaming\_reasoning\_chain}, a list of per-second reasoning entries, and (2)~\texttt{voice\_response}, a natural spoken paraphrase. We use temperature 0.3 and structured output parsing to guarantee format validity. An equivalent Chinese-language prompt is used for Chinese data, with identical constraints and structure.

\clearpage
\section{Evaluation Prompts for Reply Quality}
\label{app:eval_prompts}

We provide the three judge prompts used for the reply-quality evaluation in \S\ref{sec:reply_quality}, which is conducted via GPT-5. All prompts operate on the text transcription of the user instruction and the model response. The placeholders \texttt{\{question\}} and \texttt{\{prediction\}} are filled with the user instruction transcription and the model response transcription, respectively.

\begin{promptbox}[Evaluation Prompt: Overall Quality (Mean Score)]
       \small\ttfamily
       I need your help to evaluate the performance of several models in the speech interaction scenario. The models will receive a speech input from the user, which they need to understand and respond to with a speech output. Your task is to rate the model's responses based on the provided user input transcription [Instruction] and the model's output transcription [Response].\\[4pt]
       Please evaluate the response on a scale of 1 to 5:\\
       \quad 1 point: Largely irrelevant, incorrect, or off-topic.\\
       \quad 2 points: Somewhat relevant but lacks accuracy/completeness.\\
       \quad 3 points: Relevant and mostly accurate, may lack conciseness.\\
       \quad 4 points: Relevant, accurate, concise, clear answer.\\
       \quad 5 points: Exceptionally relevant, accurate, and to the point.\\[4pt]
       \#\#\# [Instruction]: \{question\}\\
       \#\#\# [Response]: \{prediction\}\\[4pt]
       After evaluating, please output the score only without anything else.
\end{promptbox}

\begin{promptbox}[Evaluation Prompt: Naturalness]
       \small\ttfamily
       Help evaluate the naturalness of the model response in a speech interaction scenario. Naturalness: How much does the reply sound like natural, fluent, everyday spoken conversation (not formal written language or robotic)?\\[4pt]
       Rate the reply on a scale from 1 to 5:\\
       \quad 1: Very unnatural, robotic, completely lacks spoken quality.\\
       \quad 2: Mostly unnatural, noticeably different from real speech.\\
       \quad 3: Generally natural, minor awkwardness in phrasing.\\
       \quad 4: Natural and conversational, resembling authentic speech.\\
       \quad 5: Extremely natural, indistinguishable from spontaneous talk.\\[4pt]
       \#\#\# [Instruction]: \{question\}\\
       \#\#\# [Response]: \{prediction\}\\[4pt]
       Output: Naturalness Score: X \quad Naturalness Explanation: ...
\end{promptbox}

\begin{promptbox}[Evaluation Prompt: Factual Accuracy]
       \small\ttfamily
       Help evaluate the factual accuracy of the model response in a speech interaction scenario. Factual Accuracy: Is all information mentioned factual and correct, with no hallucinations or made-up/incorrect content?\\[4pt]
       Rate the reply on a scale from 1 to 5:\\
       \quad 1: Clearly incorrect, contains hallucinations or false info.\\
       \quad 2: Partially correct, notable errors or hallucinations.\\
       \quad 3: Mostly correct and factual, minor inaccuracies.\\
       \quad 4: Accurate, no hallucinations, maybe minor details to refine.\\
       \quad 5: Fully factual, every detail accurate, no hallucination.\\[4pt]
       \#\#\# [Instruction]: \{question\}\\
       \#\#\# [Response]: \{prediction\}\\[4pt]
       Output: Accuracy Score: X \quad Accuracy Explanation: ...
\end{promptbox}

\end{document}